# Use of the Triangular Fuzzy Numbers for Student Assessment (Revised)

**Michael Gr. Voskoglou**

Department of Mathematical Sciences, School of Technological Applications,
Graduate Technological Educational Institute (T. E. I.) of Western Greece,
Patras, Greece
E-mail:  mvosk@hol.gr

## Abstract

In an earlier work we have used the Triangular Fuzzy Numbers (TFNs) as an assessment tool of student skills. This approach led to an approximate linguistic characterization of the students' overall performance, but it was not proved to be sufficient in all cases for comparing the performance of two different student groups, since two TFNs are not always comparable. In the present paper we complete the above fuzzy assessment approach by presenting a defuzzification method of TFNS based on the Center of Gravity (COG) technique, which enables the required comparison. In addition we extend our results by using the Trapezoidal Fuzzy Numbers (TpFNs) too, which are a generalization of the TFNs, for student assessment and we present suitable examples illustrating our new results in practice.

## 1. Introduction

The social demand of classifying students according to their qualifications makes the assessment of student skills a very important educational task. *Fuzzy logic*, due to its nature of characterizing the ambiguous cases with multiple values, offers rich resources for the assessment purposes. This gave us several times in past the impulse to apply principles of fuzzy logic for assessing human skills using as tools the *corresponding system's uncertainty* (e.g. see [8] and its relevant references, Section 2 of [11], etc), the *COG defuzzification technique* (e.g. see [10-11], etc), as well as two recently developed variations of the COG technique, i.e. the *Triangular (TFAM) and Trapezoidal (TpFAM) Fuzzy Assessment Models* (e.g. see [6] and [11] respectively, etc). It is of worth to notice that the TFAM and TpFAM, which are equivalent to each other (since they obtain exactly the same final results) treat better than the COG technique the ambiguous assessment cases being at the boundaries between two successive assessment grades. The use of the COG technique for assessment purposes, as well as the above mentioned two variations of it were initiated by Igor Subbotin (see [5, 6, 11], etc) Professor of Mathematics at State University in Los Angeles and Voskoglou's coauthor in many publications (e.g. [6], [11], etc).
In a recently published paper [12] we have extend our above researches by using the *Triangular Fuzzy Numbers* (*TFNs*) as an assessment tool of student skills. This approach, while it is better than our older fuzzy methods for the *individual* *assessment* [9], in case of *group assessment* led (in [12]) to an approximate characterization of the group's overall performance and *it was not proved to be always sufficient for comparing the performance of two different groups* (for more details see Example 1 of Section 4 below).

In the present paper we complete the above fuzzy assessment approach by presenting *a defuzzification method of TFNS* based on the Center of Gravity (COG) technique, which enables the required comparison of the performance of two (or more) groups. Further, we extend our results by using the *Trapezoidal Fuzzy Numbers* (*TpFNs*) too, which are generalizations of TFNs) for student assessment. The rest of the paper is organized as follows: In Section 2 we recall in brief some definitions from [12] and we present some new definition about TpFNs, which are necessary for the understanding of the paper. In Section 3 we present the defuzzification methods for TFNs/TpFNs with the COG technique. In Section 4 we consider examples illustrating our new results in practice. Finally, Section 5 is devoted to our conclusion and a brief discussion on the perspectives of future research on the subject.

## 2. Introductory Definitions

In this Section we recall in brief some definitions from Section 3.1 of [12] and we present some new definitions about TpFNs, which are necessary for the understanding of the present paper. For general facts on *fuzzy sets* we refer to the book of Klir and Folger [4].

We start with the definition of a fuzzy number:

**Definition 1:** A *Fuzzy Number* is a normal (i.e. there exists x in **R**, such that m(x) = 1) and convex (i.e. its x-cuts[*] $A^x$ are ordinary closed real intervals, for all x in [0, 1]) fuzzy set A on the set **R** of real numbers with a piecewise continuous membership function *y = m(x)*.

The following statement defines a *partial order* on the set of all FNs:

**Definition 2:** Given the FNs A and B we write A $\leq$ B (or $\geq$) if, and only if, $A_l^x \leq B_l^x$ and $A_r^x \leq B_r^x$ (or $\geq$) for all x in [0, 1]. Two FNs for which the above relation holds are called *comparable*, otherwise they are called *non comparable*.

FNs play a fundamental role in fuzzy mathematics, analogous to the role played by the ordinary numbers in classical mathematics. For general facts on FNs we refer to Chapter 3 of the book [7], which is written in Greek language, and also to the classical on the subject book [3].

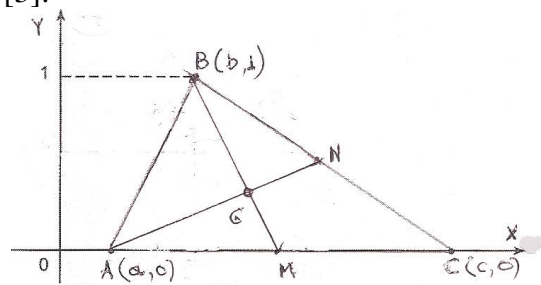

**Figure 1:** *Graph and COG of the TFN (a, b, c)*

---

[*] Let x be a real number of the interval [0, 1]. We recall then that the x-*cut* of a fuzzy set A on U, denoted by $A^x$, is defined to be the crisp set $A^x = \{y \in U: m(y) \geq x\}$.

The simplest form of FNs is probably the *Triangular FNs (TFNs)*. Roughly speaking a TFN (*a, b, c*), with *a, b* and *c* real numbers means "*approximately equal to b*" or, if you prefer, that "*b lies in the interval [a, c]*". The graph of the TFN (*a, b, c*) in the interval [*a, c*] is the union of two straight line segments forming a triangle with the X-axis, while it is zero outside [*a, c*] (see Figure 1). Therefore the analytic definition of a TFN is given as follows:

**Definition 3:** Let *a, b* and *c* be real numbers with $a < b < c$. Then the *Triangular Fuzzy Number (TFN)* A = (*a, b, c*) is the FN with membership function:

$$y = m(x) = \begin{cases} \dfrac{x-a}{b-a}, & x \in [a,b] \\ \dfrac{c-x}{c-b}, & x \in [b,c] \\ 0, & x < a \text{ and } x > c \end{cases}$$

Obviously we have that *m(b)=1*, while *b* need not be in the "middle" of *a* and *c*.

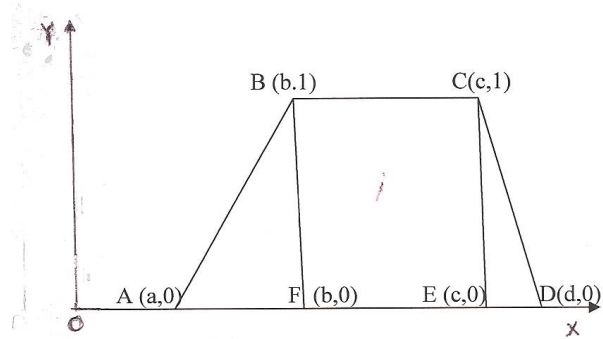

**Figure 2:** *Graph of the TpFN (a, b, c, d)*

A TpFN (*a, b, c, d*) with *a, b, c, d* in **R** actually means "*approximately in the interval [b, c]*". Its membership function *y=m(x)* is constantly 0 outside the interval [*a, d*], while its graph in this interval [*a, d*] is the union of three straight line segments forming a trapezoid with the X-axis (see Figure 2), Therefore, its analytic definition is given as follows:

**Definition 4:** Let $a < b < c < d$ be given real numbers. Then the TpFN (*a, b, c, d*) is the FN with membership function:

$$y = m(x) = \begin{cases} \dfrac{x-a}{b-a}, & x \in [a,b] \\ x = 1, & x \in [b,c] \\ \dfrac{d-x}{d-c}, & x \in [c,d] \\ 0, & x < a \text{ and } x > d \end{cases}$$

The *TpFNs are actually generalizations of TFNs*. In fact, the TFN (*a, b, d*) can be considered as a special case of the TpFN (*a, b, c, d*) with *b=c*.

It can be shown that the two well known general methods for performing operations between FNs (e.g. see Section 3 of [12]) lead to the following simple rules for the *addition* and *subtraction* of TpFNs (the same rules hold also for the TFNs; see Section 3.2 of [12]):

**Definition 5:** Let A = ($a_1$, $a_2$, $a_3$, $a_4$) and B = ($b_1$, $b_2$, $b_3$, $b_4$) be two TFNs. Then
- The sum A + B = ($a_1+b_1$, $a_2+b_2$, $a_3+b_3$, $a_4+b_4$).
- The difference A - B = A + (-B) = ($a_1-b_4$, $a_2-b_3$, $a_3-b_2$, $a_4-b_1$), where −B = ($-b_4$, $-b_3$, $-b_2$, $-b_1$) is defined to be the *opposite* of B.

In other words, the opposite of a TpFN (TFN), as well as the sum and the difference of two TpFNs (TFNs) are also TpFNs (TFNs). On the contrary, the product and the quotient of two TpFNs (TFNs), although they are FNs, *they are not always TpFNs (TFNs)*, apart from some special cases, or in terms of suitable approximating formulas (for more details see [2] and Section 3.2 of [12]).

**Definition 6:** Let A = ($a_1$, $a_2$, $a_3$, $a_4$) be a TpFN and let $k$ be a real number. Then one can define the following two *scalar operations*:
- k + A = ($k+a_1$, $k+a_2$, $k+a_3$, $k+a_4$)
- kA = ($ka_1$, $ka_2$, $ka_3$, $ka_4$), if k>0 and kA = ($ka_4$, $ka_3$, $ka_2$, $ka_1$), if k<0.

The same scalar operations can be also defined with TFNs (see Sextion 3.2of [12]).

We close this section with the following definition, which will be used Section 4 for assessing the overall performance of a human group with the help of TpFNs/TFNs:

**Definition 7:** Let $A_i$ = ($a_{1i}$, $a_{2i}$, $a_{3i}$, $a_{4i}$), i = 1, 2,…, $n$ be TpFNs/TFNs, where $n$ is a non negative integer, $n \geq 2$. Then we define the *mean value* of the $A_i$'s to be the TpFN/TFN: A = $\frac{1}{n}$ ($A_1 + A_2 + …. + A_n$).

## 3. Defuzzification of TFNs/TpFNs

In this section we shall use the COG technique for defuzzifying a given TFN/TpFN. We start with the case of TFNs:

**Proposition 1:** The coordinates (X, Y) of the COG of the graph of the TFN (*a*, *b*, *c*) are calculated by the formulas $X = \frac{a+b+c}{3}$, $Y = \frac{1}{3}$.

*Proof:* The graph of the TFN (*a*, *b*, *c*) is the triangle ABC of Figure 1, where A (*a*, 0), B (*b*, 1) and C (*c*, 0). Then, the COG, say G, of ABC is the intersection point of its medians AN and BM, where N ($\frac{b+c}{2}$, $\frac{1}{2}$) and M ($\frac{a+c}{2}$, 0). Therefore the equation of the straight line on which AN lies is $\frac{x-a}{\frac{b+c}{2}-a} = \frac{y}{\frac{1}{2}}$, or *x + (2a - b - c)y = a*  (1).

In the same way one finds that the equation of the straight line on which BM lies is
2*x* + (*a* + *c* - 2*b*)*y* = *a* + *c*   (2).

Since D = $\begin{vmatrix} 2 & a+c-2b \\ 1 & 2a-b-c \end{vmatrix} = 3(a-c) \neq 0$, the linear system of (1) and (2) has a unique solution with the respect to the variables *x* and *y* determining the coordinates of the triangle's COG.

The proof of the Proposition is completed by observing that

$$D_x = \begin{vmatrix} a+c & a+c-2b \\ a & 2a-b-c \end{vmatrix} = a^2 - c^2 + ba - bc = (a+c)(a-c) + b(a-c)$$

$$= (a-c)(a+c+b) \text{ and } D_y = \begin{vmatrix} 2 & a+c \\ 1 & a \end{vmatrix} = a - c.$$

Next, Proposition 1 will be used as a *Lemma* for the defuzzification of TpFNs. The corresponding result is the following:

***Proposition 2:*** The coordinates ($X$, $Y$) of the COG of the graph of the TpFN ($a$, $b$, $c$, $d$) are calculated by the formulas $X = \dfrac{c^2 + d^2 - a^2 - b^2 + dc - ba}{3(c+d-a-b)}$, $Y = \dfrac{2c+d-a-2b}{3(c+d-a-b)}$.

*Proof:* We divide the trapezoid forming the graph of the TpFN ($a$, $b$, $c$, $d$) in three parts, two triangles and one rectangle (Figure 2). The coordinates of the three vertices of the triangle ABE are ($a$, 0), ($b$, 1) and ($b$, 0) respectively, therefore by Proposition 1 the COG of this triangle is the point $C_1$ ($\dfrac{a+2b}{3}$, $\dfrac{1}{3}$). Similarly one finds that the COG of the triangle FCD is the point $C_2$ ($\dfrac{d+2c}{3}$, $\dfrac{1}{3}$). Also, it is easy to check that the COG of the rectangle BCFE is the point $C_3$ ($\dfrac{b+c}{2}$, $\dfrac{1}{2}$). Further, the areas of the two triangles are equal to $S_1 = \dfrac{b-a}{2}$ and $S_2 = \dfrac{d-c}{2}$ respectively, while the area of the rectangle is equal to $S_3 = c - b$.

It is well known then (e.g. see [35]) that the coordinates of the COG of the trapezoid, being the resultant of the COGs $C_i$ ($x_i$, $y_i$), for i=1, 2, 3, are calculated by the formulas $X = \dfrac{1}{S}\sum_{i=1}^{3} S_i x_i$, $Y = \dfrac{1}{S}\sum_{i=1}^{3} S_i y_i$ (3), where $S = S_1 + S_2 + S_3 = \dfrac{c+d-b-a}{2}$ is the area of the trapezoid.

The proof of the Proposition is completed by replacing the above found values of $S$, $S_i$, $x_i$ and $y_i$, i = 1, 2, 3, in formulas (3) and by performing the corresponding operations.

## 4. Use of the TFNs/TpFNs for Assessing Student Skills

We reconsider first the following example originally presented in [12]:

***Example 1:*** The students of two different Departments of the School of Management and Economics of the Graduate Technological Educational Institute (T. E. I.) of Western Greece achieved the following scores (in a climax from 0 to 100) at their common progress exam in the course "Mathematics for Economists I":

*First Department* ($D_1$): 100(2 times), 99(3), 98(5), 95(8), 94(7), 93(1), 92 (6), 90(5), 89(3), 88(7), 85(13), 82(6), 80(14), 79(8), 78(6), 76(3), 75(3), 74(3), 73(1), 72(5), 70(4), 68(2), 63(2), 60(3), 59(5), 58(1), 57(2), 56(3), 55(4), 54(2), 53(1), 52(2), 51(2), 50(8), 48(7), 45(8), 42(1), 40(3), 35(1).

*Second Department* ($D_2$) : 100(1), 99(2), 98(3), 97(4), 95(9), 92(4), 91(2), 90(3), 88(6), 85(26), 82(18), 80(29), 78(11), 75(32), 70(17), 64(12), 60(16), 58(19), 56(3), 55(6), 50(17), 45(9), 40(6).

The student performance was characterized by the *fuzzy linguistic labels (grades)* A, B, C, D and F corresponding to the above scores as follows: A (85-100) = excellent,

B (84-75) = very good, C (74-60) = good, D (59-50) = fair and F (<50) = unsatisfactory. How one can assess their individual and overall performances using the TFNs?

For this, in [12] we have assigned to each linguistic grade a TFN (denoted by the same letter) as follows: A= (85, 92.5, 100), B = (75, 79.5, 84), C = (60, 67, 74), D= (50, 54.5, 59) and F = (0, 24.5, 49). The middle entry of each of the above TFNs is equal to the mean value of the student scores assigned to the corresponding linguist grade, while its left and right entries are the minimal and maximal values of these scores respectively. In this way a TFN corresponds to each student assessing his (her) *individual* performance. Next in [12], we have calculated the mean values (see Definition 7) of the TFNs of all students of each Department (denoted by the same letters $D_1$ and $D_2$ respectively), which are:

$D_1 = \frac{1}{170} \cdot (60A+40B+20C+30D+20F) \approx (63.53, 71.74, 83.47)$ and

$D_2 = \frac{1}{255} \cdot (60A+90B+45C+45D+15F) \approx (65.88, 72.63, 79.53)$.

Observing the left entries (63.53 and 65.88 respectively) and the right entries (83.47 and 79.53 respectively) of the TFNs $D_1$ and $D_2$ one concludes that the overall performance of the two Departments could be characterized from good (C) to very good (B). However, as we have shown in Section 4.3 of [12], the above two TFNs *are non comparable* (see Definition 2), which means that it is not possible to compare the overall performance of the two Departments directly from them. Consequently, *a complementary action is needed*, in order to obtain the required comparison.

But, before of this, it is of worth to clarify that the middle entries of $D_1$ and $D_2$ (71.74 and 72.63 respectively) give *a rough approximation only* of each Department's overall performance. For this, we observe that these values *do not measure the mean performances* of the two Departments. In fact, calculating the means of the student scores in the classical way one finds the values 72.44 and 72.04 respectively, demonstrating a slightly better mean performance for $D_1$. Further, since the middle entries of the TFNs A, B, C, D and F were chosen to be equal to the means of the scores assigned to the corresponding linguistic grades, the middle entries of the TFNS $D_1$ and $D_2$ are *equal to the mean values of these means*, which justifies completely the characterization "rough approximation" given to them.

A good way to overcome this difficulty is to *defuzzify* the TFNs $D_1$ and $D_2$. By Proposition 1, the COGs of the triangles forming the graphs of the TFNs $D_1$ and $D_2$ have x-coordinates equal to $X = \frac{63.53 + 71.74 + 83.47}{3} \approx 72.91$ and

$X' = \frac{65.88 + 72.63 + 79.53}{3} \approx 72.68$ respectively.

Observe now that the GOGs of the graphs of $D_1$ and $D_2$ lie in a rectangle with sides of length 100 units on the X-axis (student scores from 0 to 100) and one unit on the Y-axis (normal fuzzy sets). Therefore, *the nearer the x-coordinate of the COG to 100, the better the corresponding Department's performance*, Thus, since $X > X'$, $D_1$ demonstrates a better overall performance than $D_2$.

Our next Example gives the opportunity of using the TpFNs too for student assessment.

*Example 2:* Six different mathematics teachers train a group of five students of the Upper Secondary Education, who won at the final stage of their National

Mathematical Competition, in order to participate in the International Mathematical Olympiad. In a preparatory test during their training the students ranked with the following scores (from 0-100) by their teachers: $S_1$ (Student 1): 43, 48, 49, 49, 50, 52, $S_2$: 81, 83. 85, 88, 91, 95, $S_3$: 76, 82, 89, 95, 95, 98, $S_4$: 86, 86, 87, 87, 87, 88 and $S_5$: 35, 40, 44, 52, 59, 62.

Assess the student performance with the help of TFNs and TpFNs.

a) *Use of the TFNs:* We consider again the TFNs A, B, C, D and F defined in Example 1. Observing the 5*6 = 30 in total student scores one finds that in the present Example we have 14 TFNs equal to A, 4 equal to B, 1 equal to C, 4 equal to D and 7 TFNs equal to F characterizing the student performance. The mean value of the above TFNs (Definition 7) is equal to M = $\frac{1}{30}$(14A + 4B + C + 4D + 7F) $\approx$ (60.33, 68.98, 79.63). Therefore, the student overall performance lies in the interval [60.33, 79.63], i.e. it could be characterized from good (C) to very good (B). Further, a rough approximation of this performance is given by the score 68.98 (good)

b) *Use of the TpFNs:* We assign to each student $S_i$ a TpFN (denoted, for simplicity, with the same letter) as follows: $S_1$ = (0, 43, 52, 59), $S_2$ = (75, 81, 95, 100), $S_3$ = (75, 76, 98, 100), $S_4$ = (85, 86, 88, 100) and $S_5$ = (0, 35, 62, 74). Each of the above TpFNs characterizes the *individual performance* of the corresponding student in the form (*a, b, c, d*), where *a* and *d* are the minimal and maximal scores respectively of the linguistic grades characterizing his/her performance, while *b* and *c* are the lower and higher scores respectively assigned to the student by the teachers.

Next, for assessing the overall student performance with the help of the above TpFNs, we calculate the mean value of the TpFNs $S_i$, i =1, 2, 3, 4, 5 (Definition 7), which is equal to the TpFN $S = \frac{1}{5}\sum_{i=1}^{5} S_i = (47, 64.2, 79, 86.6)$.

The TbFN **S** gives us the information that the student overall performance lies in the interval [*b, c*] = [64.2, 79], i.e. it could be characterized from good (C) to very good (B).

Our last Example extends the previous one giving the opportunity to apply the COG deffuzzification technique for TpFNs in order to compare the performances of two different student groups.

*Example 3:* Reconsider Example 2 assuming further that the same six teachers marked also the papers of a second group of five students (the substitutes of the previous group) examined on the same test. Assume further that the overall performance of the second group was assessed as in Example 5.2 using the TPFNs and that the mean value of the corresponding TpFNs was found to be equal to **S´** = (47.8, 65.3, 78.1, 85.9). Compare the performances of the two student groups.

For this, applying Proposition 2 one finds that the x-coordinate of the COC of the trapezoid constituting the graph of the TpFN **S** is equal to

$$X = \frac{79^2 + (86.6)^2 - (64.2)^2 - 47^2 + 79*86.6 - 47*(64.2)}{3(79 + 86.6 - 47 - 64.2)} \approx 68.84.$$

In the same way one finds that the x-coordinate of the graph of **S´** is approximately equal to 68.13. Therefore, using the same argument as that at the end of Example 1, one finds that the first group demonstrates a better overall performance.

*Remark:* In the same way as in Example 3 one can defuzzify the TpFNs $S_i$, i=1, 2, 3, 4, 5 of Example 2 corresponding to the five students of the first group. In this way it becomes possible to compare the individual performance *of any two students*, in contrast to our method presented in [9] and the equivalent to it method of A. Jones [1] that define a *partial order only* on the individual student performances.

## 5. Conclusion

In the present paper we used the TFNs/TpFNs as a tool for assessing student skills. The main advantage of the use of the TpFNs for student assessment is that in case of *individual assessment* it is sufficient for comparing the performances of *all students*, in contrast to the alternative fuzzy assessment methods applied in earlier works, which define *a partial order* only on the individual performances. However, in case of *group assessment* the TFNs/TpFNs approach *initially leads to an approximate characterization of the group's overall performance, which is not always sufficient for comparing the performances of two different groups*, as our fuzzy assessment methods applied in earlier works do. This is due to the fact that the inequality between TFNs/TpFNs defines on them a relation of partial order only. Therefore, in cases where our fuzzy outputs are not comparable, *some extra calculations are needed* in order to obtain the required comparison by defuzzifying these outputs. This could be considered as a disadvantage of this approach, although the extra calculations needed are very simple.

Our new method of using TFNs/TpFNs for the assessment of human skills is of general character, which means that it could be utilized for assessing a great variety of human (or machine; e.g. CBR systems [11]) activities. This is one of the main targets of our future research on the subject.

**Acknowledgements:** The author wishes to thank his colleagues Prof. J. Theodorou, Graduate Technological Educational Institute of Central Greece and Prof. I. Subbotin, State University, Los Angeles, for their helpful ideas about FNs.